\documentclass{article}
\usepackage[utf8]{inputenc}
\usepackage{subscript}
\usepackage{authblk}
\usepackage{setspace}
\usepackage[margin=1.25in]{geometry}
\usepackage{graphicx}
\graphicspath{ {./figures/} }
\usepackage{subcaption}
\usepackage{amsmath}
\usepackage{lineno}
\usepackage{hyperref}
\usepackage{multirow}
\usepackage{float}

\usepackage[style=authoryear, 
citestyle=authoryear,
sorting=none]{biblatex}
\let\cite\parencite
\title{Detection of trade in products derived from threatened species using machine learning and a smartphone application}
\author[1*$\dag$]{Ritwik Kulkarni}
\author[2]{WU Hanqin}
\author[1,3,4]{Enrico Di Minin}

\affil[1]{Helsinki Lab of Interdisciplinary Conservation Science, Department of Geosciences and
Geography, FI-00014, University of Helsinki, Finland}
\affil[2]{International Fund for Animal Welfare (IFAW)}
\affil[3]{Helsinki Institute of Sustainability Science (HELSUS), University of 
Helsinki, Helsinki, Finland}
 \affil[4]{School of Life Sciences, University of KwaZulu-Natal, Durban 4041, South Africa}
\date{}

\begin{document}

\maketitle

\begin{abstract}
Illegal and unsustainable wildlife trade remains a major conservation challenge, increasingly shifting to digital marketplaces where monitoring is difficult. To address this gap, we developed a deep-learning object detection framework to identify products derived from three highly trafficked species; elephant, pangolin, and tiger; using real-world images from online platforms, enforcement seizures, and market inspections. Our dataset of 39,941 images included both wildlife products and visually similar non-wildlife items, reflecting real enforcement complexity. We trained Faster R-CNN models using cross-entropy and asymmetric focal loss functions, and compared species-specific and multi-species training strategies. The best-performing model, trained with cross-entropy on all species combined, achieved an overall accuracy of 84.2\%, with high performance for pangolin (93.6\%) and tiger (90.3\%) and moderate performance for elephant (71.1\%). While asymmetric focal loss increased recall, it substantially reduced precision, limiting its utility for enforcement. We deployed the optimized model in a prototype smartphone application, potentially usable a in real-world settings for monitoring online images. On an independent test set, the app achieved 91.3\% accuracy and an F-score of 0.86. These results demonstrate the potential of object detection models to support scalable, real-time monitoring of wildlife trade, enabling rapid triage of suspicious posts and strengthening evidence-based enforcement, regulatory compliance, and conservation decision-making.
\end{abstract}

\section{INTRODUCTION}
The unsustainable trade in wild animals and plants is one of the biggest threats to biodiversity worldwide \cite{tSasRolfes2019}. Wildlife trade has been embedded in human history for millennia, with evidence of economic exchange for wildlife and their products, tracing back to early Greek and Egyptian cultures \cite{Pires2016}. However, the scale and intensity of this trade have increased dramatically in modern times, driven by the industrial revolution and rapid technological innovation \cite{Hung2014}, leading to significant declines and even extinctions of species, with far-reaching negative effects on food webs and ecosystem functioning. Thousands of species and derived products are traded for several reasons, including as pets, ornaments, trophies, medicines, food, and fashion \cite{ThomasWalters2020}. Unsustainable trade in wildlife can also threaten the livelihoods of rural communities and can pose significant risks to global food security \cite{tSasRolfes2019}. It also raises serious concerns for animal welfare and public health, particularly in relation to the spillover of zoonotic diseases \cite{Nijman2021}. 

Wildlife trade continues to expand and is also gaining ground on the Internet, where wildlife and derived products are increasingly offered for sale both legally and illegally  \cite{Davies2022} \cite{Lavorgna2014}; \cite{Morcatty2021}; \cite{Mozer2023}; \cite{SorianoRedondo2023}. This growing trend poses an even more prominent threat to biodiversity and people. The shift of wildlife trade to online and social media platforms has greatly complicated enforcement efforts \cite{Davies2022} \cite{DiMinin2018}. Sellers use websites, e-commerce, and social media to promote their products, connect with potential buyers, and negotiate prices \cite{Harrington2021}. Online trade makes it difficult to distinguish between legal and illegal activities, as national boundaries become unclear and identifying the source and destination countries is challenging \cite{Magliocca2021}. Unlike physical markets, where transactions can be observed and traced, online trade provides anonymity and a global reach, allowing traders to operate across borders with minimal risk of detection  \cite{ClementInglis2022VietnamVirtual} \cite{Lavorgna2014}. 

\begin{figure}[h]
    \centering
    \includegraphics[scale=0.2]{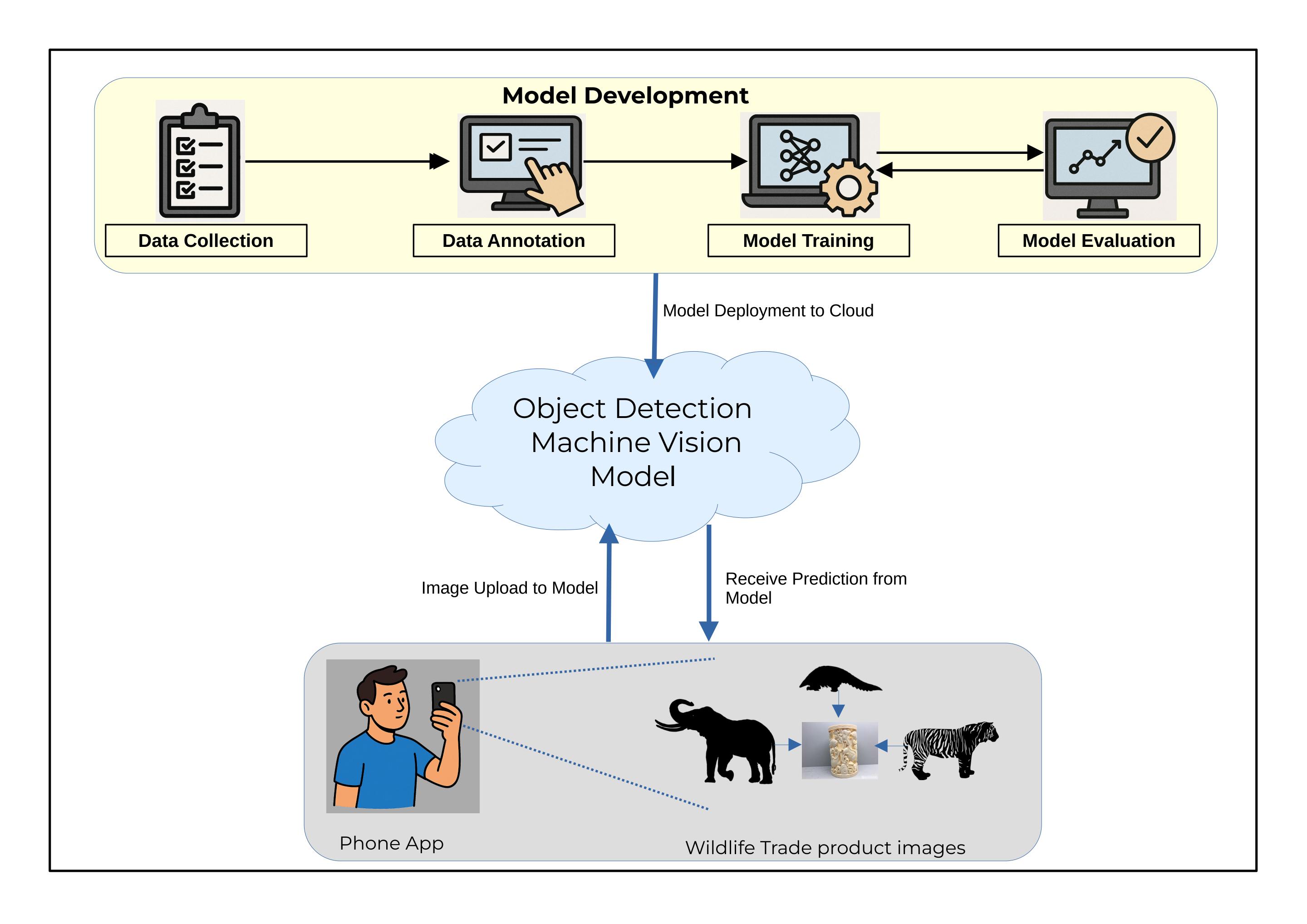}
    \caption{Machine learning based object recognition models were trained on images of elephant ivory and skins, pangolin scales and claws, and tiger skins and bones. The optimized model achieved high detection accuracy and was hosted on the cloud to be integrated into a smartphone application. This enabled real-time identification of potentially illegal products. This pipeline supports both online monitoring of wildlife trade and on-the-ground enforcement by authorities.}
    \label{fig:overview}
\end{figure}

With limited resources available to counteract the global biodiversity crisis, digital surveillance methods offer cost-effective solutions to help monitor online wildlife trade and its legality \cite{SonrickerHansen2012}. Machine learning methods, in particular, offer cost-effective solutions to automatically identify image and/or text content related to wildlife trade on digital platforms \cite{DiMinin2018}\cite{Reynolds2025}. Methods like natural language processing analyze human language through text vectorization, while machine vision models discern characteristics in images to find pertinent details \cite{Toivonen2019}. Natural language processing algorithms can search social media, e-commerce platforms, and dark web forums for wildlife trade-related keywords \cite{Kulkarni2021}. Likewise, image recognition models, mainly using convolutional neural networks, can detect species in online images, highlighting suspicious posts for additional scrutiny \cite{Kulkarni2023} \cite{Cardoso2023}. While commonly applied in ecology for automated species identification \cite{Tuia2022}, machine vision based object recognition methods capable of automatically localizing wildlife products from digital images in the context of unsustainable and illegal wildlife trade are mostly under-explored. However, see related approaches \cite{Barone2022} for an example of automated software to identify shark species from fin images (this work uses morphometric and categorical variables on fin images), and \cite{Dunbar2021} for a software to identify turtles from images (however, this is in in the context of ecology rather than directly for wildlife trade). Also see, SEE Shell a software to perform binary classification, specifically to detect the potential existence of hawksbill turtle shell (www.seeturtles.org/see-shell-instructions).

In this study, we developed a machine learning-based approach to detect and classify wildlife products using digital images. Specifically, we implemented a Faster-Region based Convolutional Neural Network (F-RCNN) model trained \cite{ren2015faster} on a dataset comprising images of elephant, pangolin, and tiger products identified either as illegally traded or having strict regulations on trade. To optimize model performance, we compared the effectiveness of two loss functions, cross-entropy loss and asymmetric focal loss, in terms of precision, recall, and classification accuracy. Our experiments included training both species-specific models and a single model capable of detecting products from all three target species. Rather than building classification models, we developed an object recognition model to precisely localize the products and reduce the presence of noise in the data \cite{ren2015faster}. In addition, we integrated the model into a smartphone-based application, enabling real-time identification of wildlife products from images. The primary intended use of the application is to support moderators of digital platforms, NGOs, and law-enforcement personnel in monitoring online wildlife trade. In this context, users may review images associated with online advertisements, social media posts, or e-commerce listings and upload them directly from their device gallery for automated screening. The application can also support field inspections as a secondary use case, where enforcement officers capture photographs of physical items using the device camera during on-site checks. In both scenarios, the system functions as a decision-support tool designed to assist human experts by rapidly triaging images and highlighting potential wildlife products for further review, rather than replacing expert judgement.

\section{Methods}
This study followed a structured workflow to develop, evaluate, and deploy an automated system for detecting wildlife trade products in images (Figure \ref{fig:overview}). The methodological pipeline began with the collection of a large dataset of wildlife product images from multiple sources, followed by expert annotation of object classes and bounding boxes. The annotated dataset was then used to train and evaluate object detection models using a Faster R-CNN framework. Model performance was assessed using standard object detection metrics to determine the most effective training configuration. To enhance interpretability, feature visualisation techniques were applied to examine the visual cues used by the model during prediction. Finally, the trained model was deployed within a cloud-based architecture and integrated into a mobile application, enabling users to capture or upload images and receive automated predictions. The following sections describe each stage of this workflow in detail.

\subsection{Data Collection}\label{data}
Data for model training were collected by members of the International Fund for Animal Welfare (IFAW\ www.ifaw.org/uk/projects/wildlife-crime-prevention-china) by monitoring illegal wildlife trade in China between 2008 and 2023 and consisted of a total of 39,941 images. The data covered illegally traded wildlife products from three species, namely elephant, pangolin, and tiger, as shown in table \ref{tab:data dist}. Image sources included: (i) online advertisements and listings on e-commerce platforms, social media, and other publicly accessible websites; (ii) photographs taken during field-based market inspections; (iii) confiscation images obtained from publicly reported law enforcement seizures; and (iv) publicly available images retrieved via search engines. The data set reflects the visual characteristics of the advertising of illegal wildlife products commonly encountered in online trade environments. To support model robustness in realistic screening scenarios, non-wildlife classes were also included. The non-wildlife class is included to train the models to discriminate between objects that appear similar to the target product but are not of animal origin. The main challenge of the model is to discriminate between such objects. These images consisted of (i) non-wildlife products sourced from IFAW’s existing archives, (ii) objects with high potential for visual confusion (e.g., jade bracelets resembling ivory bracelets and herbal medicinal slices resembling pangolin scales), and (iii) miscellaneous wildlife trade–related images that can appear in product listings but are of non target products. This composition enables a diverse and challenging set of images that not only represents real world data but also makes model evaluation results more insightful of ``on-field" performance. Images in non-wildlife class are similar to target images in aspects such as peripheral surroundings and backgrounds in the target images (see examples in supplementary Fig SF3-SF4). 

As the first data cleaning step, we removed duplicate images appearing in both, the product category and the 'non-wildlife' category (across the entire dataset). Due to the large size of the dataset, the identification of the images to be removed was done by implementing an image processing algorithm using Convolutional Neural Networks (CNNs), with the help of the Python library ImadeDedup \cite{imagededup}. The algorithm was applied pairwise to all images, resulting in a similarity score. Images with high similarity were flagged and removed after visual inspection. Only one of the images of duplicates within the class was retained. In cases of misclassification (images identified as target and non-wildlife), experts from IFAW verified the images again to classify them correctly. See Table ST7 in the supplementary for details on corrected classes.

\begin{table}[h]
    \centering
    \begin{tabular}{|l|r|}
        \hline
        \multicolumn{2}{|c|}{\textbf{elephant}} \\ \hline
        ambiguous mammoth & 5409 \\ \hline
        antique ivory & 414 \\ \hline
        elephant skin & 22 \\ \hline
        ivory & 6893 \\ \hline
        non-widlife & 12890 \\ \hline
        \hline 
        sub-total & 25,628 \\ \hline
        \multicolumn{2}{c}{} \\ \hline
        \multicolumn{2}{|c|}{\textbf{pangolin}} \\ \hline
        claw & 226 \\ \hline
        crafted scale & 283 \\ \hline
        parched scale & 65 \\ \hline
        raw scale & 427 \\ \hline
        non-widlife & 1661 \\ \hline
        \hline
        sub-total & 2662 \\ \hline
        \multicolumn{2}{c}{} \\ \hline
        \multicolumn{2}{|c|}{\textbf{tiger}} \\ \hline
        bone ambiguous & 399 \\ \hline
        claw & 538 \\ \hline
        fang ambiguous & 1993 \\ \hline
        other & 72 \\ \hline
        non-widlife & 8649 \\ \hline
        \hline
        sub-total & 11,651 \\ \hline
    \end{tabular}
    \caption{Counts of labeled images for elephant (ivory, skin, antique ivory, and ambiguous mammoth), pangolin (raw, parched, and crafted scales, claws), and tiger (bones, claws, fangs, and other products), along with corresponding non-wildlife categories. The total number of images were 39,941. These datasets formed the basis for training and evaluation of object recognition models to detect wildlife products in digital content.}
    \label{tab:data dist}
\end{table}

\subsection{Data Annotation}
Images were first provisionally grouped according to species or product category based on contextual metadata and visual characteristics. To ensure labeling reliability, all images subsequently underwent a structured expert review process. Two IFAW analysts with extensive field experience in wildlife trade monitoring and species-level product identification, independently assessed the category assignments and corrected any misclassifications. Inter-annotator discrepancies were identified and resolved through discussion to achieve consensus labeling. 
To further enhance label reliability, images that were judged to be visually ambiguous or not confidently identifiable even by experienced reviewers, were excluded from the training dataset. This exclusion criterion was implemented to reduce label noise and minimize the risk of systematic misclassification during model training.

The categories with an "ambiguous" tag were when sellers have claimed the product to be of target category, e.g. mammoth. However, it was not possible for experts to visually distinguish whether the product is from a mammoth or from elephant. Yet from the perspective of the main goal of the model (to flag suspicious images for further assessment) it is still important to include the images in the training data in order to be categorised as such.

Every image had been assigned a single label for the type of product present in the image as listed in table \ref{tab:data dist}. However, for an object detection model, bounding box information is required to indicate the precise location of the target entity within the image \cite{ren2015faster}. Unlike simple classification, which only predicts the presence of an object class, bounding boxes additionally provide spatial localisation, enabling the model to identify where objects occur within the image. Bounding boxes contain the exact (x,y) co-ordinates of rectangles drawn within the images that define the spatial region where the target object exists. Thus, as apart from additional semantics of the image, it also helps to reduce irrelevant background information from affecting the classification decision. We implemented a Web-based tool Label Studio \cite{Label} that was hosted on the servers at CSC – IT Center for Science in Finland (csc.fi/en/). Label Studio provides an interface for data annotation and can be remotely accessed. Annotation services were provided by a commercial annotation company.

As the images were quite diverse in content (see sample image examples in supplementary SF2, SF3, and SF4), e.g., some images were collages that may or may not contain the same illegal wildlife product, while in some cases only one of the images in the collage was of a wildlife product. Some of the images were of collective product displays which may contain products from non-target species as well as non-wildlife related products. To reduce noise in the data, annotators were instructed to draw separate bounding boxes for products where a single box was clearly differentiated and marked. However, when clear segregation was not possible (e.g., a pile of various pangolin scales) larger boxes were drawn over groups of target objects.  Once annotation was completed, labelled data and associated images were further processed using image augmentation, namely horizontal flip (images are randomly flipped akin to a mirror image) and Normalisation (the RGB values of the image are normalised to span a limited range \cite{shorten2019survey}). Image augmentation is done to help achive stable training iterations of the model and increase model generalisation by exposing the model to varied representations of the same data.”

\subsection{Model Training}
We used the Faster Region-based Convolutional Neural Network (Faster R-CNN) \cite{ren2015faster} model for object detection. Faster R-CNN is a two-stage detection framework that consists of a Region Proposal Network (RPN) and a region-based detection network. The RPN generates candidate object proposals (a list of possible bounding boxes that clearly define which part within the image the target object is located), and the detection network classifies these proposals into object categories and refines their bounding-box coordinates. The integration of the RPN with the detection network allows Faster R-CNN to efficiently identify objects in an image while maintaining high accuracy. The core component where features are detected in the Faster R-CNN model is called the backbone of the model. The Densenet121  \cite{huang2017densely} architecture was used for the backbone based on results observed through the comparative evaluation of architectures for wildlife images in \cite{Kulkarni2023}. 

During the training phase, Faster R-CNN optimizes a multi-task loss that combines classification loss for object categories and regression loss for bounding box refinement. This dual optimization ensures accurate object detection and precise localization. In this study, we compared the performance of Faster R-CNN using two different loss functions for the classification task: cross-entropy loss  \cite{ridnik2021asymmetric} and asymmetric focal loss \cite{ridnik2021asymmetric}. Cross-entropy loss is a widely used loss function for classification tasks. It measures the dissimilarity between the predicted probability distribution and the true distribution. While cross-entropy loss is effective in many scenarios, it is less effective in cases of severe class imbalance \cite{ridnik2021asymmetric}. We tested the utility of this loss function as the data is skewed towards the elephant class. In object detection, the vast majority of proposals correspond to the background class, which can dominate the loss function and hinder the ability of the model to learn meaningful representations for rare object classes. To address the limitations of cross-entropy loss, we experimented using asymmetric focal loss, a variant of focal loss designed to handle extreme class imbalance more effectively. Focal loss was originally proposed to down-weight the contribution of well-classified examples, focusing the model's attention on hard examples that are difficult to classify. As the data was strongly biased towards elephant products, we also developed three separate models for each of the species to delineate the impact of size and type of data in model training. Smaller data models were tested for both cross-entropy loss and asymmetric focal loss. 

Models were trained by fine-tuning a  Densenet121 backbone architecture that was pre-trained on ImageNet data \cite{imagenet_cvpr09}. During training, the parameters of all layers of the model were updated. Training was stopped if there was no improvement in performance on the validation set for 15 consecutive epochs. The learning rate was incrementally increased and the training cycle was repeated until the stopping threshold was reached again. See Tables ST5 and ST6 in the Supplementary for a full list of parameters. 

The dataset was partitioned into training and validation subsets using an 80/20 split. To ensure representation across all product categories, the split was stratified by class so that each species–product category appeared proportionally in both subsets. Only the training subset was used for model optimization, while the validation subset was used for early stopping and hyperparameter selection.

\subsection{Model Evaluation}
Model performance was evaluated using Mean Average Precision (mAP) and Mean Average Recall (mAR) \cite{metrics}. mAP measures the precision-recall trade-off by computing the area under the precision-recall curve across different Intersection over Union (IoU) thresholds, providing an overall assessment of detection accuracy. In other words, mAP measures how accurately the model detects and classifies objects, combining both localization (how well the predicted bounding box matches the object) and classification (whether the correct species is identified). The subscript “50” (mAP\_50) indicates that a detection is considered correct when the predicted bounding box overlaps the ground truth box by at least 50\% (i.e., an Intersection-over-Union threshold of 0.50). mAR, on the other hand, quantifies the model’s ability to retrieve relevant object instances by averaging recall across multiple IoU thresholds. mAR measures how completely the model can find relevant objects across the dataset. The subscript “100” (mAR\_100) indicates that up to 100 detection proposals per image are considered when computing recall, reflecting the model’s ability to retrieve true objects among multiple candidates. Higher mAP and mAR values indicate better detection performance, balancing precision and recall effectively. We calculated mAP and mAR separately for each of the categories in the training data. However, it is important to note that in this particular case mAP and mAR may under-report the recall and precision due to the complexity of the data. See Fig SF2-panel (a) in the supplementary for an example. Due to the complex display of target objects, even though the model has correctly identified the object, it may not match the annotation perfectly in the number or span of the boxes. This is because the annotators have drawn boxes to the best of their ability to cover all objects using multiple boxes of a single category per image. Sometimes, several objects are grouped together (e.g. a collection of items scattered in the image), this necessitates that sets of multiple boxes to be drawn to cover all the items. In such cases, object boundaries detected by the model after training, although semantically correct, may differ from annotation bounding boxes. In other words, predicted bounding boxes cover the same region as annotation but using different box configurations, thus incorrectly reducing the mAP and mAR score. Despite this, due to limitations in the annotation task (impractical to draw a bounding box on all non-wildlife objects in a scene as that is an unlimited category) only single category boxes were marked. Thus, an image may have e.g. ivory in it but may also contain elements such as plastic rings or metal holdings categorised as `non wildlife'. Once trained, the model is able to mark bounding boxes from all the categories in the training data. Thus, the `non-wildlife' objects in the images are detected along with the target category, even though the ground truth annotations may have only the wildlife boxes. This also reduces the performance scores to an extent. To evaluate the model without the mentioned discrepancies, we calculated the accuracy of detecting the class  along with mAP and mAR for the models. 

\subsection{Feature Visualisation}
Gradient-weighted Class Activation Mapping (Grad-CAM) \cite{selvaraju2017grad} was employed to visualise feature representations learned by the Faster R-CNN model, providing interpretability by highlighting image regions influential in the model’s decision-making. Grad-CAM is a qualitative interpretability technique that generates heatmaps indicating which regions of an input image contribute most strongly to a model’s prediction (See supplementary for more information on Grad-CAM algorithm). This visualization helps researchers interpret the model’s behaviour by identifying the image features that the network prioritizes when forming its decision. Specifically, we selected the last layer of the backbone network as the target layer for gradient computation, since the features of this layer are used later for localisation and classification, thus ensuring a high-level semantic understanding of detected objects. While Grad-CAM is widely used for convolutional neural networks (CNNs) in classification tasks, its application to object detection models like Faster R-CNN poses challenges due to the presence of multiple region proposals and the two-stage detection pipeline. The spatial misalignment between proposal-based feature maps and final detections complicates the direct attribution of activations, necessitating additional processing steps to obtain meaningful visualisations (See supplementary). However, despite the limitations, the method can still be very useful as illustrated in Figure \ref{fig:cam-ivory} and has no impact on final classification of the model.

\subsection{Phone App Building and Testing}
\begin{figure}[h!]
    \centering
    \includegraphics[scale=0.1]{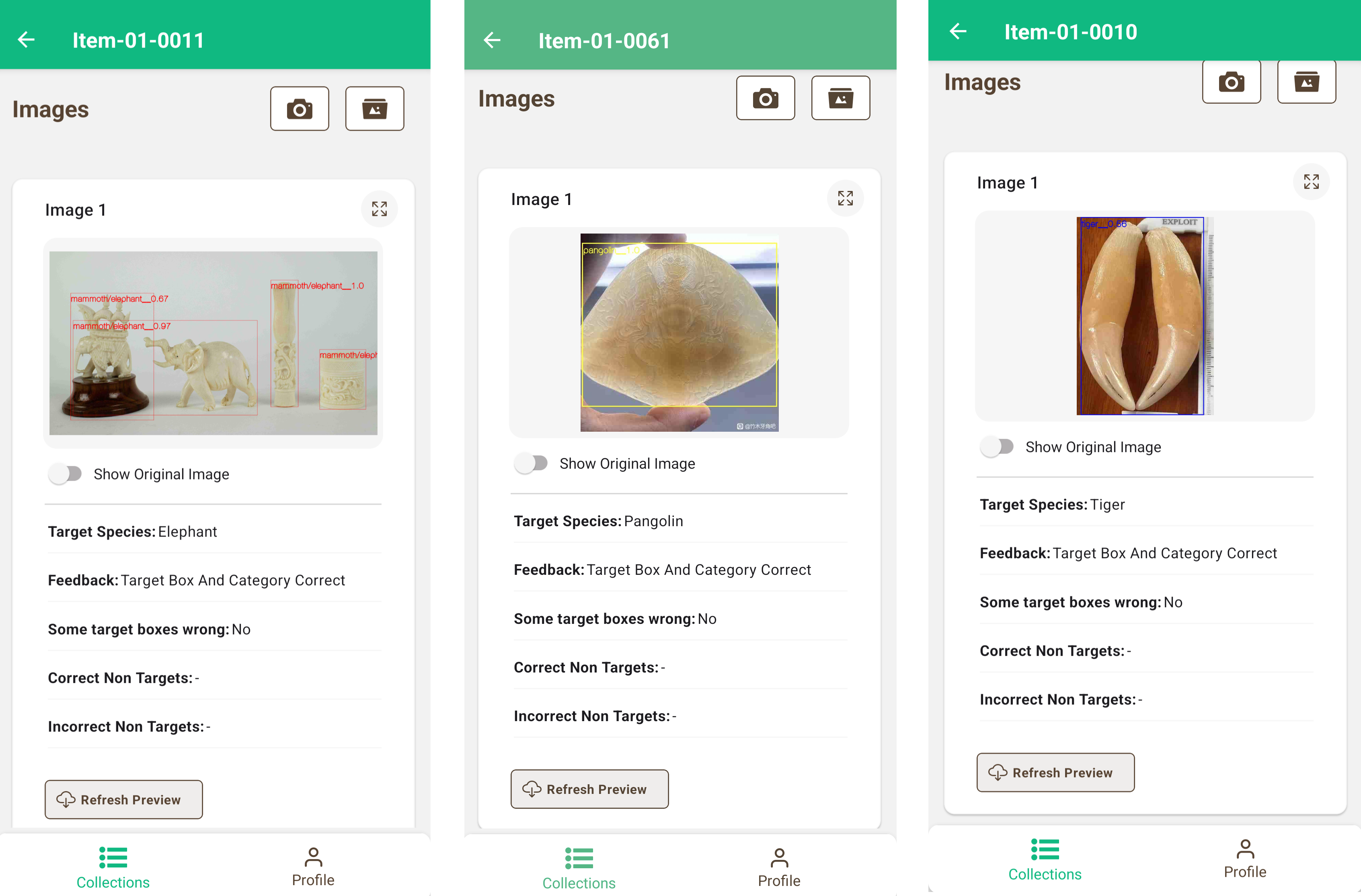}
    \caption{Screenshot of the app depicting the prediction of detected ivory, crafted pangolin scale and ambiguous tiger fang products. The app also has the ability to collect feedback on the results of the prediction.}
    \label{fig:phone}
\end{figure}

To demonstrate the end usability of the machine learning model beyond the academic exercise of training and testing the model. A mobile application was developed (Figure \ref{fig:phone}) that provided a user-friendly interface to interact with the machine learning model and have the ability to get prediction on individual images. The object detection mobile application was developed using React Native (reactnative.dev/). The app integrates React Native Vision Camera to capture high-quality images efficiently. The backend was built with Java \cite{arnold2005java} and Spring Boot \cite{walls2015spring}, while MongoDB \cite{chauhan2019review} serves as the database for managing the data. This technology stack ensures a robust and efficient system as these components enable rapid image capture, stable communication between the app and the model, and efficient storage and retrieval of detection results, which are essential for a system intended for real-time use by operational stakeholders. Since this also follows standard software development architecture, stakeholders can develop this design further for their use case. 

The application workflow was designed to be simple and intuitive for operational users. Upon launching the application, users are first presented with a login interface that ensures secure access to the system. After authentication, the main dashboard displays a list of previously captured and processed images, allowing users to review past detections and associated results. From this interface, users may also initiate the collection of new data. When capturing new images, the application provides two options: selecting an existing image from the device gallery or capturing a new image directly using the device camera. Once an image is selected or captured, it is uploaded to the cloud-based backend where the machine learning model performs object detection and returns predictions in the form of bounding boxes and associated class labels. Users can adjust the sensitivity threshold of the model predictions to control the level of detection confidence displayed. The interface additionally supports expert feedback on the model outputs, enabling users to evaluate the correctness of both the predicted bounding boxes and the assigned classifications. This feedback mechanism facilitates human-in-the-loop evaluation and provides a pathway for iterative improvement and validation of the deployed model. (See video in supplementary)  

To test the usability and performance of the App we sourced an additional set of images (n=138) for the products of the three species in the dataset. The additional images were collected in the same manner as described in section \ref{data}. The app was designed to take feedback on the accuracy of the prediction and store it in the database. We calculated the overall accuracy of the App based on the performance of the model on these images. 

\section{Results}
The following results present the performance metrics of a Faster R-CNN (F-RCNN) model trained for the classification of wildlife trade images, evaluated using two loss functions: cross-entropy loss and asymmetric focal loss. Model performance is assessed using mean Average Precision (mAP) at an IoU threshold of 0.50 (mAP\_50), mean Average Recall (mAR\_100), category-wise mAP and mAR, and classification accuracy for four categories: 'elephant', 'tiger', 'pangolin', and 'non-wildlife'.

\begin{figure}[h]
    \centering
    \includegraphics[scale=0.4]{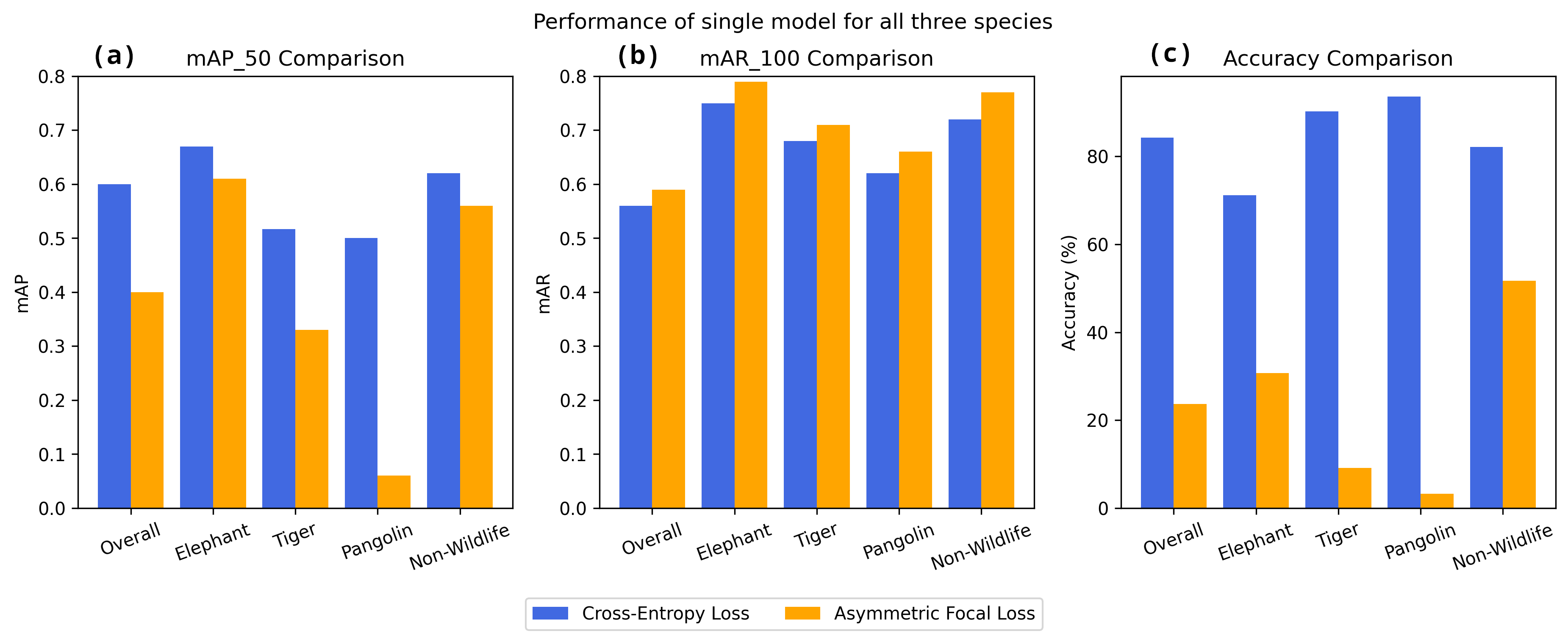}
    \caption{Performance comparison of multi-species model trained across all three species. Bar chart illustrating model performance using two loss functions; cross-entropy loss and asymmetric focal loss, on a combined dataset of elephant, tiger, pangolin, and non-wildlife categories. Metrics include mean average precision at 50\% IoU (mAP\_50), mean average recall at 100 detections (mAR\_100), category-wise mAP and mAR, and overall accuracy. Under cross-entropy loss, the model achieved higher performance, with overall mAP\_50 of 0.60, mAR\_100 of 0.56, and species-level accuracies of 71.1\% (elephant), 90.3\% (tiger), 93.6\% (pangolin), and 82.1\% (non-wildlife). In contrast, asymmetric focal loss produced lower accuracies, 30.7\%, 9.2\%, 3.3\%, and 51.7\% for elephant, tiger, pangolin, and non-wildlife, respectively, despite slightly higher recall for some categories. The visualization highlights the relative robustness of cross-entropy loss for detecting wildlife products across diverse taxa and underscores the challenge of optimizing precision–recall trade-offs when developing a single model for multi-species detection.}
    \label{fig:single-model_bar}
\end{figure}

\clearpage

As shown in Figure \ref{fig:single-model_bar} (see also tables ST1-ST4 in the Supplementary materials), when trained with cross-entropy loss, the model achieved an overall mAP\_50 of 0.60 and an overall mAR\_100 of 0.56. The category-wise mAP values indicate the highest precision for elephant (0.67) and the lowest for pangolin (0.50), while mAR values suggest the highest recall for elephant (0.75) and the lowest for pangolin (0.62). The classification accuracy per category reveals a strong performance for pangolin (93.6\%) and tiger (90.3\%), whereas the elephant and non-wildlife classes achieved 71.1\% and 82.1\%, respectively. In contrast, asymmetric focal loss caused a significant drop in overall mAP\_50 (0.40) and a marginal improvement in overall mAR\_100 (0.59). The category-wise mAP scores indicate a marked reduction, particularly for pangolin (0.06) and tiger (0.33). Similarly, the accuracy for these categories decreased substantially, with pangolin achieving only 3.3\% and tiger 9.2\%. However, the recall values remained relatively stable across all categories, with elephant achieving the highest mAR (0.79) and pangolin the lowest (0.66). These results suggest that while cross-entropy loss yields better precision and classification accuracy, asymmetric focal loss improves recall, but at the cost of precision and overall model reliability, particularly for under-represented classes such as pangolin.

\begin{figure}[H]
    \centering
    \includegraphics[scale=0.5]{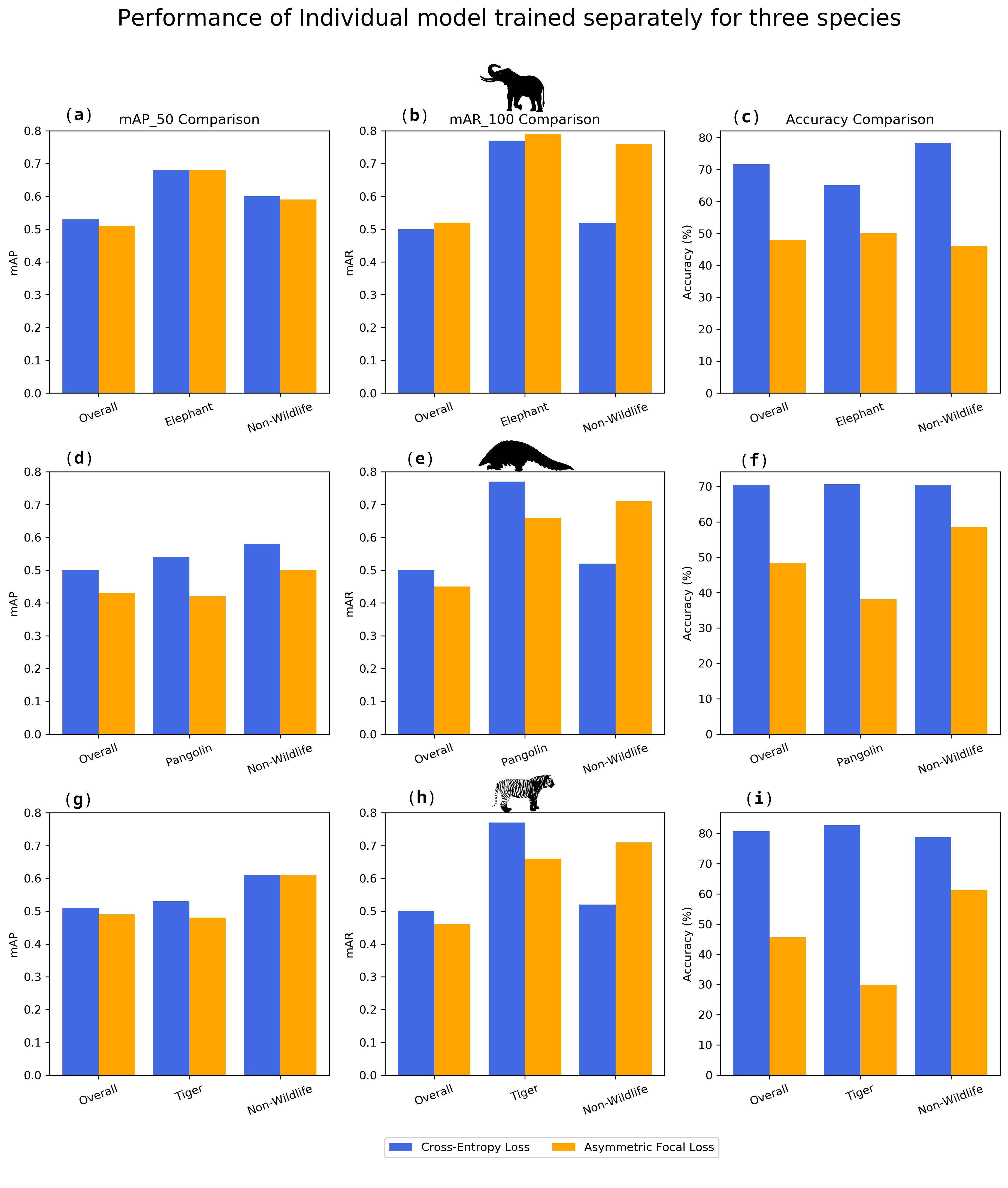}
    \caption{Bar charts summarize the performance of individual models trained on single-species datasets under two loss functions; cross-entropy loss and asymmetric focal loss. Each row panel represents one species (elephant, pangolin, and tiger), with results shown for species products versus non-wildlife images. Performance metrics include mean average precision at 50\% IoU (mAP\_50), mean average recall at 100 detections (mAR\_100), category-wise mAP and mAR, and classification accuracy. Across species, cross-entropy consistently produced higher accuracies—65.0\% for elephant products, 70.6\% for pangolin products, and 82.7\% for tiger products, compared to substantially lower accuracies under asymmetric focal loss (50.0\%, 38.1\%, and 29.9\%, respectively). While focal loss achieved comparable or slightly higher recall in some categories, it generally reduced precision and overall accuracy. Multi-species model (Figure \ref{fig:single-model_bar}) showed better performance across most of the metrics compared to per species models. }
    \label{fig:indv_models}
\end{figure}

When comparing the performance of the models trained separately for each species (Figure \ref{fig:indv_models}) against the single model trained on all three species together (Figure \ref{fig:single-model_bar}), several key trends emerge. The model trained on all species together using cross-entropy loss achieved a higher overall mAP\_50 (0.60) compared to individual models, where elephant (0.53), pangolin (0.50), and tiger (0.51) all exhibited lower mAP\_50 values. This suggests that training on a diverse dataset may enhance generalization and improve precision. Similarly, for mAR\_100, which measures recall, the individual models for elephant (0.50), pangolin (0.45), and tiger model (0.45) performed slightly worse than the all-species model (0.56). Notably, category-wise analysis reveals that the elephant product model had a category-wise mAP (0.68) comparable to the all-species model (0.67), indicating that species with larger dataset representation may benefit less from specialization. Conversely, the pangolin class exhibited a lower category-wise mAP (0.50 vs. 0.54) for the multi-species model, suggesting that rarer species might not benefit from multi-species training as much as common ones for box localisation. The Asymmetric Focal Loss (AFL) consistently underperformed in mAP\_50 across all models, reinforcing its tendency to prioritize recall over precision. Interestingly, the per-class mAP for elephant (0.68), pangolin (0.54) and tiger (0.53) in the individual models exceeded that in the all-species model 0.67, 0.50, and 0.51, respectively), suggesting that training separate models may improve bounding box localisation precision for certain species. However, per class accuracies suffered when trained separately, indicating that the presence of multiple species in training may help in feature learning for under-represented classes. Overall, while the all-species model demonstrates better accuracy and generalization, individual models may offer improved box localisation for specific species, particularly for well-represented classes like elephant and tiger.

\begin{figure}[H]
    \centering
    \includegraphics[scale=0.2]{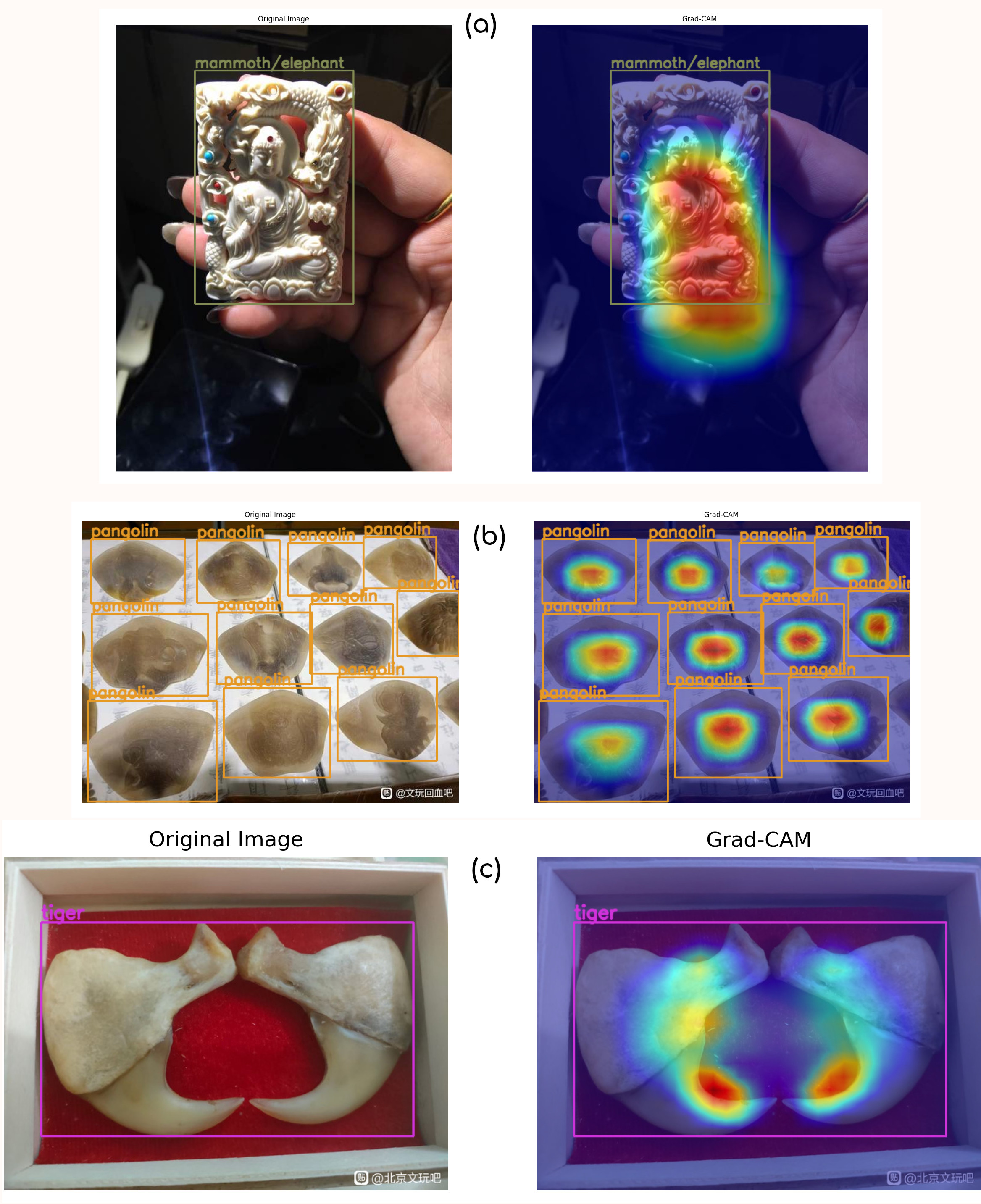}
    \caption{Gradient-weighted Class Activation Mapping (Grad-CAM) outputs for Faster R-CNN models applied to (a) carved elephant ivory, (b) pangolin scales, and (c) tiger claws. Heatmaps highlight key discriminative features: intricate carvings and textures in ivory, localized scale patterns in pangolin, and sharp curved edges in tiger claws. These activations illustrate how the model identifies species-specific visual cues for classification.}
    \label{fig:cam-ivory}
\end{figure}

The Gradient-weighted Class Activation Mapping (Grad-CAM) visualizations correspond to
Faster R-CNN object detection outputs for three distinct wildlife products: tiger claws, pangolin
scales, and a carved elephant ivory artifact. The first image (Figure 5.a , depicting an ivory carving,
shows high activation along the carved details of the object, with some diffusion into the surrounding
hand. This suggests that the model is leveraging intricate carvings and surface texture as key features
for localising the product and predominantly basing decision making on features in these regions rather than outside the product. These features can be then used in distinguishing ivory artifacts. However, it is important to note that the carvings themselves are not an indicative feature of ivory (see additional examples in the supplementary figures SF3-SF4). The second image 5.b, representing pangolin scales, demonstrates a well-localized and consistent Grad-CAM response, with strong activations centred on each identified scale. This indicates that the model correctly associates key discriminative features such as
shape and texture with the pangolin classification. Figure \ref{fig:cam-ivory}. represents tiger claws, and similarly exhibits a pattern, with the Grad-CAM heatmap predominantly highlighting the sharper edges of the claws and along the curved shape of the claw and bone. These Grad-CAM visualizations provide
insight into the interpretability and reliability of Faster R-CNN in detecting illegal wildlife products,
highlighting potential areas for refinement, particularly in cases of mislocalized or inverted heatmaps.

In certain cases (see Figure SF1 in the supplementary), Gradient-weighted Class Activation Mapping (Grad-CAM) visualizations exhibit an inverted activation pattern, where low activation appears over the target object and high activation is observed in the surrounding background. This phenomenon is particularly noticeable in complex models such as Faster R-CNN, where the interplay between region proposals, feature extraction, and classification layers may lead to non-intuitive activation distributions. Despite this inversion, the model’s predictions, including both classification and bounding box localization, remain accurate, suggesting that the network utilizes background context as an auxiliary cue for decision-making. This behaviour highlights the complexity of feature attribution in deep neural networks and underscores the need for cautious interpretation of saliency maps in object detection models.

\begin{table}[h!]
    \centering
    \begin{tabular}{|c|c|c|c|c|c|c|c|}
    \hline
        \textbf{Species} & \textbf{TP}  & \textbf{FP}  & \textbf{FN} & \textbf{Precision} & \textbf{Recall}   & \textbf{F-Score} & \textbf{Sensitivity} \\
        \hline
         Elephant & 38 & 15 & 1 & 0.72 & 0.97 & 0.83 &  97.4\%  \\
         \hline
         Tiger & 35 & 2 & 4 & 0.95 & 0.89 & 0.92 & 89.7\% \\
         \hline
         Pangolin & 38 & 9  & 1 & 0.80 & 0.97 & 0.88 &  97.4\% \\
         \hline
         Non Wildlife & 15 & 3 & 6 & 0.83 & 0.71 & 0.76 & 71.4\% \\
         \hline
         \hline
         Overall & 126 & 29 & 12 & 0.81 & 0.91 & 0.86 & 91.3\% \\
         \hline
    \end{tabular}
    \caption{Real-world performance of the deployed smartphone application.
Performance metrics of the trained model when served through a mobile app and tested on live images of elephant, tiger, and pangolin products, as well as non-wildlife items. TP stands for True Positive, FP - False Positive, and FN False Negative. The app achieved high sensitivity for elephant (97.4\%), pangolin (97.4\%), and tiger (89.7\%) products, with lower performance on non-wildlife images (71.4\%). Overall sensitivity (same as total accuracy) was 91.3\%, demonstrating the model’s effectiveness for real-time detection in practical settings. In the same order F-score was 0.83, 0.92, 0.88, 0.76 and 0.88. }
    \label{tab:app_performance}
\end{table}

The performance evaluation of the model deployed via the phone app shows a strong classification performance across key target species. As shown in Table \ref{tab:app_performance}, the model achieved high detection rate (sensitivity) for elephant and pangolin classes (97.4\% each), with a comparatively lower but still decent performance for Tiger (89.7\%). The Non-Wildlife class, representing potentially ambiguous non-target items, showed reduced sensitivity at 71.4\%, indicating room for improvement in distinguishing non-trade items from wildlife products. Overall, the model attained a sensitivity of 91.3\% across 138 user-submitted test cases (by definition this is the same as total model accuracy). Error analysis  revealed that among the 29 false positive instances, the majority were misidentified as elephant (15 times) and pangolin (9 times), suggesting a possible over-representation or bias toward these dominant classes. Similarly, there were 12 false negatives with "Non-Wildlife" having the highest at 6, followed by tiger at 4, 1 each for elephant and pangolin.

\section{Discussion}

The findings of this study demonstrate that deep learning based object detection has the potential to substantially enhance of conservation practitioners, enforcement agencies, and online platform moderators to identify wildlife products in digital environments \cite{Stringham2021} \cite{Kulkarni2023}. The Faster R‑CNN model trained with cross‑entropy loss provided strong overall performance particularly in precision and classification sensitivity, supporting its practical applicability in large‑scale monitoring systems designed to counter illegal wildlife trade. Although asymmetric focal loss offered some improvements in recall, its substantial reduction in accuracy for under‑represented classes underscores the importance of selecting loss functions that balance sensitivity with operational reliability. For conservation applications, minimising false positives is critical for preventing wasted enforcement effort, while avoiding false negatives reduces the risk of illegal products going undetected. 

A  strength of this study is the demonstration that a single multi‑species detection model can use visual features from other taxa and outperform species‑specific models across several metrics. This finding is highly relevant for conservation science and practice, where monitoring programs often span multiple taxa and must detect a wide diversity of products under variable conditions. By learning shared visual features across species, the multi‑species model benefits from a richer, more heterogeneous dataset, supporting better generalization to new imagery and emerging  patterns. This is particularly valuable for conservation contexts where novel wildlife products or product presentations may appear abruptly in markets or online spaces. In contrast, the species-specific models showed slightly improved classification sensitivity (mAR) for elephant and pangolin, suggesting that species‑level models may still be useful in targeted enforcement scenarios or specialized monitoring programs, provided significant training data is available.

Some limitations observed, particularly in detecting elephant products, highlight both methodological challenges and conservation-relevant complexities. Ivory carvings often share visual characteristics with legally traded substitutes such as bone, resin, or mammoth ivory, and sellers frequently exploit this ambiguity to avoid detection \cite{SharmaEtAl2025IvoryOrBoneATRFTIR}  \cite{SantosEtAl2025MammothTaskIsotopes}. Machine vision models may partially rely on contextual cues rather than product-specific features when distinguishing such items \cite{Kulkarni2023}. From a conservation science perspective, this highlights the importance of developing models that better capture species-specific morphological traits and that incorporate complementary data streams (e.g., text metadata, seller behaviour analyses, platform patterns) to improve decision-making \cite{Cardoso2023}.

The successful deployment of the model in a smartphone application represents a significant step toward operationalizing AI for conservation. Law enforcement officers, customs officials, and market inspectors often work under time constraints and with limited access to specialized expertise  \cite{LynamEtAl2025RisingTideConservationTechnology}. The ability to capture images and receive real-time predictions increases the speed and efficiency of field inspections, allowing frontline practitioners to triage suspicious items rapidly \cite{LynamEtAl2025RisingTideConservationTechnology}. The model’s high sensitivity for elephant and pangolin products, two of the world’s most heavily trafficked mammals, enhances its potential to support global conservation priorities, including actions aligned with CITES enforcement and national wildlife crime strategies.

In digital contexts, where the volume of online trade is rapidly expanding, the model can help conservation organizations and online platforms automatically flag high‑risk listings for further human review. This is increasingly important as online trade imagery is often intentionally obscured or manipulated to evade detection. Continuous monitoring is essential because wildlife traffickers adapt their presentation strategies in response to enforcement pressure, creating non-stationary data that can degrade model performance over time \cite{UNODC2024WWCRPressRelease}. Periodic retraining with updated datasets, including newly confiscated products or emerging trade trends, will therefore be needed to maintain effectiveness \cite{DongEtAl2024DataDriftPVLDB}.

From a broader conservation science perspective, automated detection systems such as the one presented here can significantly strengthen wildlife crime prevention efforts by improving situational awareness across trade networks. They can assist in identifying emerging trafficking routes, temporal shifts in demand, and species at increasing risk, all of which are critical factors for designing targeted conservation interventions. Such tools should not replace expert judgement, but rather function as force multipliers that expand monitoring capacity, allowing experts to focus on more complex investigative tasks.

Future refinements will further enhance conservation utility. Strategies such as model pruning, quantization, and lightweight architectures could improve deployment in low‑resource settings, expanding access to conservation practitioners globally \cite{DeepCompression}. Integration of uncertainty estimation and confidence scoring would support more informed decision-making, especially in high-stakes enforcement scenarios \cite{Nallapareddy2023}. Expanding annotated datasets to include a larger diversity of species and crafted wildlife products would extend applicability to a wider range of conservation concerns. Developing complementary text‑analysis models could also enhance detection of wildlife trade where images are absent or ambiguous.

Overall, this study demonstrates the strong potential of AI‑driven object detection to transform the monitoring and regulation of wildlife trade. By enabling scalable, accurate, and accessible detection of illegal products across both digital and physical markets, these tools can help conservation practitioners respond more rapidly to emerging threats, allocate limited enforcement resources more effectively, and support global efforts to protect species targeted by the illegal wildlife trade.





\


\subsection*{Data Availability}
Scripts for model and app related functions are available in research data repository at Etsin (http://urn.fi/urn:nbn:fi:fd-d5f60f55-d474-3275-8e49-fb77cfb93488)

\printbibliography

@misc{Label,
  title={{Label Studio}: Data labeling software},
  url={https://github.com/heartexlabs/label-studio},
  note={Open source software available from https://github.com/heartexlabs/label-studio},
  author={
    Maxim Tkachenko and
    Mikhail Malyuk and
    Andrey Holmanyuk and
    Nikolai Liubimov},
  year={2020-2022},
}

@article{Kulkarni2023,
  title = {Towards automatic detection of wildlife trade using machine vision models},
  volume = {279},
  ISSN = {0006-3207},
  url = {http://dx.doi.org/10.1016/j.biocon.2023.109924},
  DOI = {10.1016/j.biocon.2023.109924},
  journal = {Biological Conservation},
  publisher = {Elsevier BV},
  author = {Kulkarni,  Ritwik and Di Minin,  Enrico},
  year = {2023},
  month = mar,
  pages = {109924}
}

@misc{imagededup,
  title={Imagededup},
  author={Tanuj Jain and Christopher Lennan and Zubin John and Dat Tran},
  year={2019},
  howpublished={\url{https://github.com/idealo/imagededup}},
}

@article{ren2015faster,
  title={Faster r-cnn: Towards real-time object detection with region proposal networks},
  author={Ren, Shaoqing and He, Kaiming and Girshick, Ross and Sun, Jian},
  journal={Advances in neural information processing systems},
  volume={28},
  year={2015},
  url={https://proceedings.neurips.cc/paper_files/paper/2015/file/14bfa6bb14875e45bba028a21ed38046-Paper.pdf}
}

@inproceedings{huang2017densely,
  title={Densely connected convolutional networks},
  author={Huang, Gao and Liu, Zhuang and Van Der Maaten, Laurens and Weinberger, Kilian Q},
  booktitle={Proceedings of the IEEE conference on computer vision and pattern recognition},
  pages={4700--4708},
  year={2017},
  url={openaccess.thecvf.com/content_cvpr_2017/papers/Huang_Densely_Connected_Convolutional_CVPR_2017_paper.pdf}
}

@inproceedings{ridnik2021asymmetric,
  title={Asymmetric loss for multi-label classification},
  author={Ridnik, Tal and Ben-Baruch, Emanuel and Zamir, Nadav and Noy, Asaf and Friedman, Itamar and Protter, Matan and Zelnik-Manor, Lihi},
  booktitle={Proceedings of the IEEE/CVF international conference on computer vision},
  url={https://openaccess.thecvf.com/content/ICCV2021/papers/Ridnik_Asymmetric_Loss_for_Multi-Label_Classification_ICCV_2021_paper.pdf},
  pages={82--91},
  year={2021}
}

@inproceedings{imagenet_cvpr09,
        AUTHOR = {Deng, J. and Dong, W. and Socher, R. and Li, L.-J. and Li, K. and Fei-Fei, L.},
        TITLE = {{ImageNet: A Large-Scale Hierarchical Image Database}},
        BOOKTITLE = {CVPR09},
        YEAR = {2009},
        BIBSOURCE = "http://www.image-net.org/papers/imagenet_cvpr09.bib"}

@article{metrics,
  title = {A Comparative Analysis of Object Detection Metrics with a Companion Open-Source Toolkit},
  volume = {10},
  ISSN = {2079-9292},
  url = {http://dx.doi.org/10.3390/electronics10030279},
  DOI = {10.3390/electronics10030279},
  number = {3},
  journal = {Electronics},
  publisher = {MDPI AG},
  author = {Padilla,  Rafael and Passos,  Wesley L. and Dias,  Thadeu L. B. and Netto,  Sergio L. and da Silva,  Eduardo A. B.},
  year = {2021},
  month = jan,
  pages = {279}
}

@inproceedings{selvaraju2017grad,
  title={Grad-cam: Visual explanations from deep networks via gradient-based localization},
  author={Selvaraju, Ramprasaath R and Cogswell, Michael and Das, Abhishek and Vedantam, Ramakrishna and Parikh, Devi and Batra, Dhruv},
  booktitle={Proceedings of the IEEE international conference on computer vision},
  pages={618--626},
  year={2017},
  url = {https://openaccess.thecvf.com/content_ICCV_2017/papers/Selvaraju_Grad-CAM_Visual_Explanations_ICCV_2017_paper.pdf}
}

@book{arnold2005java,
  title={The Java programming language},
  author={Arnold, Ken and Gosling, James and Holmes, David},
  year={2005},
  publisher={Addison Wesley Professional}
}

@book{walls2015spring,
  title={Spring Boot in action},
  author={Walls, Craig},
  year={2015},
  publisher={Simon and Schuster}
}

@article{chauhan2019review,
  title={A review on various aspects of MongoDB databases},
  author={Chauhan, Anjali},
  journal={International Journal of Engineering Research \& Technology (IJERT)},
  volume={8},
  number={05},
  pages={90--92},
  year={2019},
  DOI = {10.17577/IJERTV8IS050031},
  url = {https://www.ijert.org/a-review-on-various-aspects-of-mongodb-databases}
}

@inbook{Pires2016,
  title = {The Illegal Wildlife Trade},
  ISBN = {9780199935383},
  url = {http://dx.doi.org/10.1093/oxfordhb/9780199935383.013.161},
  DOI = {10.1093/oxfordhb/9780199935383.013.161},
  booktitle = {Oxford Handbooks Online: Criminology and Criminal Justice},
  publisher = {Oxford University Press},
  author = {Pires,  Stephen F. and Moreto,  William D.},
  year = {2016},
  month = jul 
}

@article{Hung2014,
  title = {Drastic population fluctuations explain the rapid extinction of the passenger pigeon},
  volume = {111},
  ISSN = {1091-6490},
  url = {http://dx.doi.org/10.1073/pnas.1401526111},
  DOI = {10.1073/pnas.1401526111},
  number = {29},
  journal = {Proceedings of the National Academy of Sciences},
  publisher = {Proceedings of the National Academy of Sciences},
  author = {Hung,  Chih-Ming and Shaner,  Pei-Jen L. and Zink,  Robert M. and Liu,  Wei-Chung and Chu,  Te-Chin and Huang,  Wen-San and Li,  Shou-Hsien},
  year = {2014},
  month = jun,
  pages = {10636–10641}
}

@article{ThomasWalters2020,
  title = {Motivations for the use and consumption of wildlife products},
  volume = {35},
  ISSN = {1523-1739},
  url = {http://dx.doi.org/10.1111/cobi.13578},
  DOI = {10.1111/cobi.13578},
  number = {2},
  journal = {Conservation Biology},
  publisher = {Wiley},
  author = {Thomas‐Walters,  Laura and Hinsley,  Amy and Bergin,  Daniel and Burgess,  Gayle and Doughty,  Hunter and Eppel,  Sara and MacFarlane,  Douglas and Meijer,  Wander and Lee,  Tien Ming and Phelps,  Jacob and Smith,  Robert J. and Wan,  Anita K. Y. and Veríssimo,  Diogo},
  year = {2020},
  month = aug,
  pages = {483–491}
}

@article{tSasRolfes2019,
  title = {Illegal Wildlife Trade: Scale,  Processes,  and Governance},
  volume = {44},
  ISSN = {1545-2050},
  url = {http://dx.doi.org/10.1146/annurev-environ-101718-033253},
  DOI = {10.1146/annurev-environ-101718-033253},
  number = {1},
  journal = {Annual Review of Environment and Resources},
  publisher = {Annual Reviews},
  author = {‘t Sas-Rolfes,  Michael and Challender,  Daniel W.S. and Hinsley,  Amy and Veríssimo,  Diogo and Milner-Gulland,  E.J.},
  year = {2019},
  month = oct,
  pages = {201–228}
}

@article{Nijman2021,
  title = {Covid-19,  biodiversity conservation and welfare of wild animals partially under human control},
  volume = {10},
  ISSN = {1800-427X},
  url = {http://dx.doi.org/10.47605/tapro.v10i1.243},
  DOI = {10.47605/tapro.v10i1.243},
  number = {1},
  journal = {Taprobanica},
  publisher = {Research Center for Climate Change,  University of Indonesia},
  author = {Nijman,  Vincent},
  year = {2021},
  month = may,
  pages = {1–3}
}

@article{Lavorgna2014,
  title = {Wildlife trafficking in the Internet age},
  volume = {3},
  ISSN = {2193-7680},
  url = {http://dx.doi.org/10.1186/s40163-014-0005-2},
  DOI = {10.1186/s40163-014-0005-2},
  number = {1},
  journal = {Crime Science},
  publisher = {Springer Science and Business Media LLC},
  author = {Lavorgna,  Anita},
  year = {2014},
  month = may 
}

@article{DiMinin2018,
  title = {A framework for investigating illegal wildlife trade on social media with machine learning},
  volume = {33},
  ISSN = {1523-1739},
  url = {http://dx.doi.org/10.1111/cobi.13104},
  DOI = {10.1111/cobi.13104},
  number = {1},
  journal = {Conservation Biology},
  publisher = {Wiley},
  author = {Di Minin,  Enrico and Fink,  Christoph and Hiippala,  Tuomo and Tenkanen,  Henrikki},
  year = {2018},
  month = nov,
  pages = {210–213}
}

@article{Harrington2021,
  title = {Live wild animal exports to supply the exotic pet trade: A case study from Togo using publicly available social media data},
  volume = {3},
  ISSN = {2578-4854},
  url = {http://dx.doi.org/10.1111/csp2.430},
  DOI = {10.1111/csp2.430},
  number = {7},
  journal = {Conservation Science and Practice},
  publisher = {Wiley},
  author = {Harrington,  Lauren A. and Auliya,  Mark and Eckman,  Harry and Harrington,  Alix P. and Macdonald,  David W. and D’Cruze,  Neil},
  year = {2021},
  month = apr 
}

@article{Magliocca2021,
  title = {Comparative Analysis of Illicit Supply Network Structure and Operations: Cocaine,  Wildlife,  and Sand},
  volume = {3},
  ISSN = {2516-7227},
  url = {http://dx.doi.org/10.31389/jied.76},
  DOI = {10.31389/jied.76},
  number = {1},
  journal = {Journal of Illicit Economies and Development},
  publisher = {LSE Press},
  author = {Magliocca,  Nicholas and Torres,  Aurora and Margulies,  Jared and McSweeney,  Kendra and Arroyo-Quiroz,  Inés and Carter,  Neil and Curtin,  Kevin and Easter,  Tara and Gore,  Meredith and H\"{u}bschle,  Annette and Massé,  Francis and Rege,  Aunshul and Tellman,  Elizabeth},
  year = {2021},
  pages = {50–73}
}

@article{SonrickerHansen2012,
  title = {Digital Surveillance: A Novel Approach to Monitoring the Illegal Wildlife Trade},
  volume = {7},
  ISSN = {1932-6203},
  url = {http://dx.doi.org/10.1371/journal.pone.0051156},
  DOI = {10.1371/journal.pone.0051156},
  number = {12},
  journal = {PLoS ONE},
  publisher = {Public Library of Science (PLoS)},
  author = {Sonricker Hansen,  Amy L. and Li,  Annie and Joly,  Damien and Mekaru,  Sumiko and Brownstein,  John S.},
  editor = {Johnson,  Christopher James},
  year = {2012},
  month = dec,
  pages = {e51156}
}

@article{Reynolds2025,
  title = {The potential for AI to revolutionize conservation: a horizon scan},
  volume = {40},
  ISSN = {0169-5347},
  url = {http://dx.doi.org/10.1016/j.tree.2024.11.013},
  DOI = {10.1016/j.tree.2024.11.013},
  number = {2},
  journal = {Trends in Ecology \& Evolution},
  publisher = {Elsevier BV},
  author = {Reynolds,  Sam A. and Beery,  Sara and Burgess,  Neil and Burgman,  Mark and Butchart,  Stuart H.M. and Cooke,  Steven J. and Coomes,  David and Danielsen,  Finn and Di Minin,  Enrico and Durán,  América Paz and Gassert,  Francis and Hinsley,  Amy and Jaffer,  Sadiq and Jones,  Julia P.G. and Li,  Binbin V. and Mac Aodha,  Oisin and Madhavapeddy,  Anil and O’Donnell,  Stephanie A.L. and Oxbury,  William M. and Peck,  Lloyd and Pettorelli,  Nathalie and Rodríguez,  Jon Paul and Shuckburgh,  Emily and Strassburg,  Bernardo and Yamashita,  Hiromi and Miao,  Zhongqi and Sutherland,  William J.},
  year = {2025},
  month = feb,
  pages = {191–207}
}

@article{Toivonen2019,
  title = {Social media data for conservation science: A methodological overview},
  volume = {233},
  ISSN = {0006-3207},
  url = {http://dx.doi.org/10.1016/j.biocon.2019.01.023},
  DOI = {10.1016/j.biocon.2019.01.023},
  journal = {Biological Conservation},
  publisher = {Elsevier BV},
  author = {Toivonen,  Tuuli and Heikinheimo,  Vuokko and Fink,  Christoph and Hausmann,  Anna and Hiippala,  Tuomo and J\"{a}rv,  Olle and Tenkanen,  Henrikki and Di Minin,  Enrico},
  year = {2019},
  month = may,
  pages = {298–315}
}

@article{Kulkarni2021,
  title = {Automated retrieval of information on threatened species from online sources using machine learning},
  volume = {12},
  ISSN = {2041-210X},
  url = {http://dx.doi.org/10.1111/2041-210X.13608},
  DOI = {10.1111/2041-210x.13608},
  number = {7},
  journal = {Methods in Ecology and Evolution},
  publisher = {Wiley},
  author = {Kulkarni,  Ritwik and Di Minin,  Enrico},
  year = {2021},
  month = may,
  pages = {1226–1239}
}

@article{Cardoso2023,
  title = {Detecting wildlife trafficking in images from online platforms: A test case using deep learning with pangolin images},
  volume = {279},
  ISSN = {0006-3207},
  url = {http://dx.doi.org/10.1016/j.biocon.2023.109905},
  DOI = {10.1016/j.biocon.2023.109905},
  journal = {Biological Conservation},
  publisher = {Elsevier BV},
  author = {Cardoso,  Ana Sofia and Bryukhova,  Sofiya and Renna,  Francesco and Reino,  Luís and Xu,  Chi and Xiao,  Zixiang and Correia,  Ricardo and Di Minin,  Enrico and Ribeiro,  Joana and Vaz,  Ana Sofia},
  year = {2023},
  month = mar,
  pages = {109905}
}

@article{Mozer2023,
  title = {An introduction to illegal wildlife trade and its effects on biodiversity and society},
  volume = {3},
  ISSN = {2666-9374},
  url = {http://dx.doi.org/10.1016/j.fsiae.2023.100064},
  DOI = {10.1016/j.fsiae.2023.100064},
  journal = {Forensic Science International: Animals and Environments},
  publisher = {Elsevier BV},
  author = {Mozer,  Annika and Prost,  Stefan},
  year = {2023},
  month = dec,
  pages = {100064}
}

@article{Morcatty2021,
  title = {Online trade in wildlife and the lack of response to COVID-19},
  volume = {193},
  ISSN = {0013-9351},
  url = {http://dx.doi.org/10.1016/j.envres.2020.110439},
  DOI = {10.1016/j.envres.2020.110439},
  journal = {Environmental Research},
  publisher = {Elsevier BV},
  author = {Morcatty,  Thais Q. and Feddema,  Kim and Nekaris,  K.A.I. and Nijman,  Vincent},
  year = {2021},
  month = feb,
  pages = {110439}
}

@article{SorianoRedondo2023,
  title = {Online wildlife trade in species of conservation concern},
  volume = {16},
  ISSN = {1755-263X},
  url = {http://dx.doi.org/10.1111/conl.12985},
  DOI = {10.1111/conl.12985},
  number = {6},
  journal = {Conservation Letters},
  publisher = {Wiley},
  author = {Soriano‐Redondo,  Andrea and Alwasiti,  Haider and Kulkarni,  Ritwik and Correia,  Ricardo A. and Bryukhova,  Sofiya and Lita,  Nermalie M. and Rigor,  Lodigario A. and Tejerero,  Darwin R. and Tenazas,  Theresa M. and Di Minin,  Enrico},
  year = {2023},
  month = oct 
}

@article{Barone2022,
  title = {Performance of iSharkFin in the identification of wet dorsal fins from priority shark species},
  volume = {68},
  ISSN = {1574-9541},
  url = {http://dx.doi.org/10.1016/j.ecoinf.2021.101514},
  DOI = {10.1016/j.ecoinf.2021.101514},
  journal = {Ecological Informatics},
  publisher = {Elsevier BV},
  author = {Barone,  Monica and Mollen,  Frederik H. and Giles,  Jenny L. and Marshall,  Lindsay J. and Villate-Moreno,  Melany and Mazzoldi,  Carlotta and Pérez-Costas,  Elisa and Heine,  J\"{u}rgen and Guisande,  Cástor},
  year = {2022},
  month = may,
  pages = {101514}
}

@article{Stringham2021,
  title = {A guide to using the internet to monitor and quantify the wildlife trade},
  volume = {35},
  ISSN = {1523-1739},
  url = {http://dx.doi.org/10.1111/cobi.13675},
  DOI = {10.1111/cobi.13675},
  number = {4},
  journal = {Conservation Biology},
  publisher = {Wiley},
  author = {Stringham,  Oliver C. and Toomes,  Adam and Kanishka,  Aurelie M. and Mitchell,  Lewis and Heinrich,  Sarah and Ross,  Joshua V. and Cassey,  Phillip},
  year = {2021},
  month = mar,
  pages = {1130–1139}
}

@article{SharmaEtAl2025IvoryOrBoneATRFTIR,
  title   = {Ivory or Bone? discrimination using ATR-FTIR spectroscopy and chemometrics},
  author  = {Sharma, Chandra Prakash and Bhatia, Dimple and Singh, Rajinder},
  journal = {Science \& Justice},
  volume  = {65},
  number  = {4},
  pages   = {101261},
  year    = {2025},
  doi     = {10.1016/j.scijus.2025.101261}
}

@article{SantosEtAl2025MammothTaskIsotopes,
  title   = {A mammoth task: stable isotope analyses as a tool to prevent illegal trade of elephant ivory},
  author  = {Santos, Maria E. A. and Toropov, Pavel and Agarwal, Pihu and Frichot, Pierre Archimede Jonathan and Tilley, Hannah Bethany and Wan, Wilson Wan Zhongyue and Chan, Jovy and Baker, David Michael},
  journal = {Frontiers in Ecology and Evolution},
  volume  = {13},
  pages   = {1533703},
  year    = {2025},
  doi     = {10.3389/fevo.2025.1533703}
}

@article{LynamEtAl2025RisingTideConservationTechnology,
  title   = {The rising tide of conservation technology: empowering the fight against poaching and unsustainable wildlife harvest},
  author  = {Lynam, Antony J. and Cronin, Drew T. and Wich, Serge A. and Steward, Jessica and Howe, Antony and Kolla, Nandini and Markovina, Marina and Torrico, Orlando and Reyes, Veronica and Sophalrachana, Kimsrann and Stevens, X and Schmidt, E and Cox, H},
  journal = {Frontiers in Ecology and Evolution},
  volume  = {13},
  pages   = {1527976},
  year    = {2025},
  doi     = {10.3389/fevo.2025.1527976}
}

@article{DongEtAl2024DataDriftPVLDB,
  title   = {Efficiently Mitigating the Impact of Data Drift on Machine Learning Pipelines},
  author  = {Dong, Sijie and Wang, Qitong and Sahri, Soror and Palpanas, Themis and Srivastava, Divesh},
  journal = {Proceedings of the VLDB Endowment},
  volume  = {17},
  number  = {11},
  pages   = {3072--3081},
  year    = {2024},
  doi     = {10.14778/3681954.3681984}
}

@misc{UNODC2024WWCRPressRelease,
  title        = {Despite two decades of concerted action, wildlife trafficking persists worldwide with more than 4,000 species affected, says new UNODC World Wildlife Crime Report},
  author       = {{United Nations Office on Drugs and Crime}},
  year         = {2024},
  month        = {May},
  note         = {Press release, Vienna, 13 May 2024},
  howpublished = {\url{https://www.unodc.org/unodc/en/press/releases/2024/May/despite-two-decades-of-concerted-action--wildlife-trafficking-persists-worldwide-with-more-than-4-000-species-affected--says-new-unodc-world-wildlife-crime-report.html}}
}

@article{Davies2022,
  title = {Identifying opportunities for expert‐mediated triangulation in monitoring wildlife trade on social media},
  volume = {36},
  ISSN = {1523-1739},
  url = {http://dx.doi.org/10.1111/cobi.13858},
  DOI = {10.1111/cobi.13858},
  number = {2},
  journal = {Conservation Biology},
  publisher = {Wiley},
  author = {Davies,  Alisa and Hinsley,  Amy and Nuno,  Ana and Martin,  Rowan O.},
  year = {2022},
  month = jan 
}

@article{Tuia2022,
  title = {Perspectives in machine learning for wildlife conservation},
  volume = {13},
  ISSN = {2041-1723},
  url = {http://dx.doi.org/10.1038/s41467-022-27980-y},
  DOI = {10.1038/s41467-022-27980-y},
  number = {1},
  journal = {Nature Communications},
  publisher = {Springer Science and Business Media LLC},
  author = {Tuia,  Devis and Kellenberger,  Benjamin and Beery,  Sara and Costelloe,  Blair R. and Zuffi,  Silvia and Risse,  Benjamin and Mathis,  Alexander and Mathis,  Mackenzie W. and van Langevelde,  Frank and Burghardt,  Tilo and Kays,  Roland and Klinck,  Holger and Wikelski,  Martin and Couzin,  Iain D. and van Horn,  Grant and Crofoot,  Margaret C. and Stewart,  Charles V. and Berger-Wolf,  Tanya},
  year = {2022},
  month = feb 
}

@techreport{ClementInglis2022VietnamVirtual,
  title       = {Vietnam’s virtual landscape for illicit wildlife trading: A snapshot of e-commerce and social media},
  author      = {Cl{\'e}ment, Th{\'e}o and Inglis, Sam},
  year        = {2022},
  month       = {10},
  type        = {Trend Report},
  institution = {Global Initiative Against Transnational Organized Crime (GI-TOC)},
  address     = {Geneva, Switzerland},
  url         = {https://read-me.org/s/GITOC-MMFU-Vietnams-virtual-landscape-for-illicit-wildlife-trading-A-snapshot-of-e-commerce-and-soci.pdf},
  note        = {Unspecified DOI (report)}
}

@article{Dunbar2021,
  title = {HotSpotter: Using a computer-driven photo-id application to identify sea turtles},
  volume = {535},
  ISSN = {0022-0981},
  url = {http://dx.doi.org/10.1016/j.jembe.2020.151490},
  DOI = {10.1016/j.jembe.2020.151490},
  journal = {Journal of Experimental Marine Biology and Ecology},
  publisher = {Elsevier BV},
  author = {Dunbar,  Stephen G. and Anger,  Edward C. and Parham,  Jason R. and Kingen,  Colin and Wright,  Marsha K. and Hayes,  Christian T. and Safi,  Shahnaj and Holmberg,  Jason and Salinas,  Lidia and Baumbach,  Dustin S.},
  year = {2021},
  month = feb,
  pages = {151490}
}

@inproceedings{Nallapareddy2023,
  title = {EvCenterNet: Uncertainty Estimation for Object Detection Using Evidential Learning},
  url = {http://dx.doi.org/10.1109/IROS55552.2023.10341826},
  DOI = {10.1109/iros55552.2023.10341826},
  booktitle = {2023 IEEE/RSJ International Conference on Intelligent Robots and Systems (IROS)},
  publisher = {IEEE},
  author = {Nallapareddy,  Monish R. and Sirohi,  Kshitij and Drews,  Paulo L. J. and Burgard,  Wolfram and Cheng,  Chih-Hong and Valada,  Abhinav},
  year = {2023},
  month = oct,
  pages = {5699–5706}
}

@misc{DeepCompression,
  doi = {10.48550/ARXIV.1510.00149},
  url = {https://arxiv.org/abs/1510.00149},
  author = {Han,  Song and Mao,  Huizi and Dally,  William J.},
  title = {Deep Compression: Compressing Deep Neural Networks with Pruning,  Trained Quantization and Huffman Coding},
  publisher = {arXiv},
  year = {2015},
  copyright = {arXiv.org perpetual,  non-exclusive license}
}

@article{shorten2019survey,
  title   = {A survey on image data augmentation for deep learning},
  author  = {Shorten, Connor and Khoshgoftaar, Taghi M.},
  journal = {Journal of Big Data},
  volume  = {6},
  number  = {1},
  pages   = {60},
  year    = {2019},
  doi     = {10.1186/s40537-019-0197-0}
}
\end{document}